\documentclass[letterpaper, 10 pt, conference]{ieeeconf}  

\usepackage{graphicx}
\usepackage{url}
\usepackage{orcidlink}
\usepackage{amsmath,amsfonts}
\DeclareMathOperator*{\argmax}{arg\,max}
\usepackage{etoolbox}
\usepackage[style=numeric,sorting=none,sortcites=true]{biblatex}
\makeatletter
\patchcmd{\@makecaption}
  {\scshape}
  {}
  {}
  {}
\makeatletter
\patchcmd{\@makecaption}
  {\\}
  {.\ }
  {}
  {}
\makeatother
\def\tablename{Table}

\IEEEoverridecommandlockouts                              

\title{\LARGE \bf
A Surprisingly Efficient Representation for Multi-Finger Grasping}

\author{ Hengxu Yan$^*$, Hao-Shu Fang$^*$, Cewu Lu
\thanks{Hengxu Yan, Hao-Shu Fang, Cewu Lu are with Shanghai Jiao Tong University, Cewu Lu is the corresponding author. Email:
        {\tt\small HengxuYan@sjtu.edu.cn, fhaoshu@gmail.com, lucewu@sjtu.edu.cn }%
}}

\begin{document}

\maketitle
\renewcommand{\thefootnote}{\fnsymbol{footnote}}
\footnotetext[1]{The first two authors contributed equally to this work.}
\thispagestyle{empty}
\pagestyle{empty}

\begin{abstract}

The problem of grasping objects using a multi-finger hand has received significant attention in recent years. However, it remains challenging to handle a large number of unfamiliar objects in real and cluttered environments. In this work, we propose a representation that can be effectively mapped to the multi-finger grasp space. Based on this representation, we develop a simple decision model that generates accurate grasp quality scores for different multi-finger grasp poses using only hundreds to thousands of training samples. We demonstrate that our representation performs well on a real robot and achieves a success rate of 78.64\% after training with only 500 real-world grasp attempts and 87\% with 4500 grasp attempts. Additionally, we achieve a success rate of 84.51\% in a dynamic human-robot handover scenario using a multi-finger hand.
\end{abstract}

\section{INTRODUCTION}

Grasping is a critical function that enables robots to interact with the physical world, and numerous scenarios require more than parallel antipodal grasping. Multi-finger grasping can further enable in-hand robotic manipulation and play a crucial role in the overall grasp problem.

So far, most methods generate multi-finger grasp pose candidates directly from raw observations, such as point clouds or RGB images, as demonstrated in~\cite{7354004, xu2023unidexgrasp, 10160982, 10160667, wan2023unidexgrasp++, Lundell2020MultiFinGANGC, jiang2021hand, turpin2022grasp}.  Specifically, \cite{Lundell2020MultiFinGANGC} proposes a methodology for estimating the pose of an object within a scene and subsequently learning an offset based on the object's pose to determine the grasp pose. Similarly, \cite{jiang2021hand, turpin2022grasp} focus on predicting the contact area between the object and a multi-finger hand. However, a limitation common to these methods is their capacity to grasp only a single object. Additionally, the multi-finger grasp pose presents significant complexities and variations due to the high degree of freedom of the multi-finger hand, making the labeling of grasp data extremely costly. Furthermore, since there are different types of multi-finger grippers available on the market, establishing a multi-finger grasping dataset for each type of gripper is costly. Consequently, achieving satisfactory grasping performance on multiple grippers is currently impractical.
\begin{figure}[!t]
\centering
\includegraphics[width=0.45\textwidth]{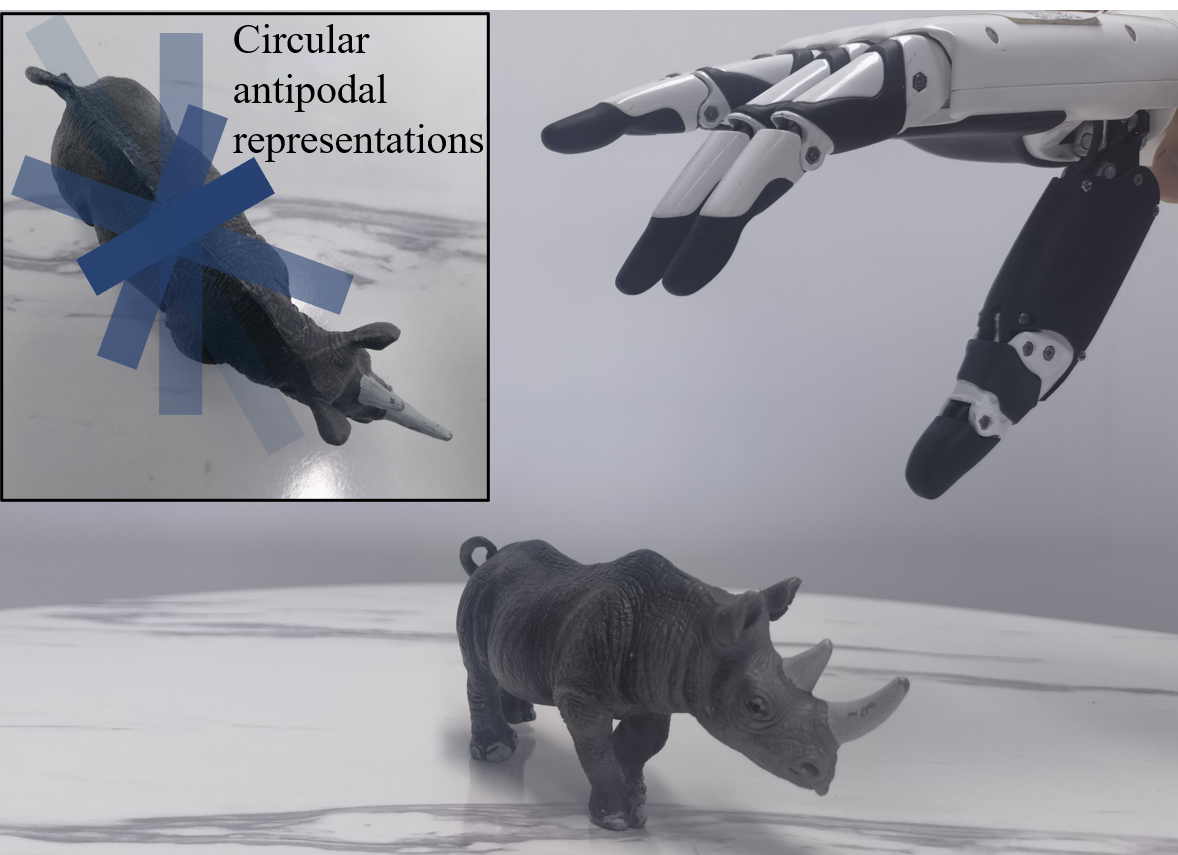}\\
\caption{The circular antipodal representations for multi-finger grasping. The less transparent the representation is, the higher the circular antipodal score.}
\label{fig_grasping}
\vspace{-0.5cm}
\end{figure}

In our study, we introduce a novel intermediate representation aimed at the generation of multi-finger grasp poses, emphasizing minimal data requirements while preserving robust performance. Our methodology is structured into two principal phases. The initial phase involves the prediction of dense circular antipodal representations distributed throughout the observed scene. These representations facilitate the sampling of multiple candidates for multi-finger grasp poses. For each designated grasp location upon an arbitrary curved surface within the scene, this representation is characterized by two key parameters: \textbf{the antipodal scores} and \textbf{the minimum antipodal grasp widths}, given different in-plane rotations and grasp depths along the grasp viewpoint. It implicitly characterizes the surface geometry of the object. Subsequently, the representation corresponding to each grasp candidate is input into a decision model, which serves the purpose of identifying the specific multi-finger grasp type. To simplify this process, we discretize the multi-finger grasp space, generating a set of feasible multi-finger grasp poses based on a predefined taxonomy of 16 grasp types. These generated poses serve as the pool from which the decision model selects an appropriate grasp type. This procedural pipeline effectively reduces the complexity of the decision space, all the while maintaining the flexibility of the multi-finger hand. To validate the efficacy of our representation design, we compare our representation with several network features, providing empirical evidence of its internal structural effectiveness. Fig.~\ref{fig_grasping} shows the circular antipodal representation for multi-finger grasping.

We conduct experiments in the real world to verify the effectiveness of the representation, including grasping static objects in cluttered scenes and reactive grasping in a handover setting. For grasping in cluttered scenes, we randomly pour diverse objects onto a plate, and a robot empties the plate with a multi-finger hand. During each grasp attempt, our method generate multiple high-quality multi-finger grasp poses for a partial-view point cloud. The best grasp pose for execution is selected by considering both collision avoidance and grasp quality. We achieve a success rate of 78.64\% after training with only 500 real-world grasp attempts and 87\% with 4500 on over 300 unseen objects of varying sizes, shapes, and materials.

For the reactive multi-finger hand-over case, an in-hand camera is attached to the robot's wrist, enabling the execution of closed-loop control. This is achieved by continuously predicting the grasp pose based on the current observation and subsequently servoing to the target pose. We show that we can achieve the temporal smoothness property for the intermediate representation, which makes it easy to grasp dynamic objects. In a handover setting, we report an 84.51\% success rate on 30 objects.

We summarize our primary contributions in the following aspects:
\begin{itemize}
    \item Identifying and developing an efficient representation tailored for multi-finger grasping.
    \item Demonstrating that our representation surpasses the performance of other unstructured network features.
    \item Achieving exceptional performance with our method, necessitating only a few hundred grasp attempts.
    \item Illustrating the temporal smoothness property of our representation and its successful application in reactive grasping scenarios.
\end{itemize}

\section{Related work}
To date, most of the research in this field has been focusing on the problem of parallel antipodal grasping, with only a small subset of studies investigating multi-finger grasping. As a result, this section will be divided into two parts, each addressing one of these two topics: multi-finger grasping and parallel antipodal grasping.
\subsection{Multi-finger Grasping}

During the early stages of multi-finger grasp research, the primary focus is on developing effective force closure strategies, as demonstrated by studies such as \cite{Forceclosure1, Forceclosure2, rodriguez2012caging, rosales2012synthesis, prattichizzo2012manipulability, liu2021synthesizing}. However, a major challenge with this approach is the requirement for an accurate model of each object in the environment \cite{6672028}. This can limit the model's applicability and make it difficult to generalize. Moreover, the overall complexity of the model can lead to significant computational costs, resulting in only a small number of feasible grasp poses being generated.

Subsequently, many learning-based approaches emerge, which can overcome these challenges and achieve greater efficiency. Due to the high cost of obtaining data through multi-finger grasping, many studies resort to collecting trial-and-errors in the simulated environment \cite{qin2022one}, where a domain gap inevitably exists. Some studies borrow demonstrations of human grasping \cite{brahmbhatt2019contactgrasp, kokic2020learning} and use model-free methods to imitate human learning, while others adopt pure reinforcement learning approaches. Additionally, some studies focus on generating efficient multi-finger hand grasp poses dataset \cite{chao2021dexycb, taheri2020grab}. For instance, \cite{wang2022dexgraspnet} proposes a large-scale dataset of 1.32 million grasp poses for 5355 objects, spanning more than 133 categories of objects. Meanwhile, \cite{zhu2021toward} constructs a dataset that can train a model to maintain contact with objects as much as possible.

Most of studies previously mentioned predominantly rely on raw perception and necessitate substantial data usage. Often, these studies are narrowly focused on grasping either a single object or a limited array of objects that are methodically arranged, thereby constraining their broader applicability. In contrast, our model, employing an intermediate representation, markedly reduces data requirements and demonstrates the capability to grasp a multitude of objects that are randomly stacked together.

\subsection{Parallel Antipodal Grasping}

Parallel antipodal grasping involves generating multiple grasp poses for a given scene. Early research focuses on 2D planes and predicts 4 degrees of freedom, as seen in studies such as \cite{traditional1, traditional2, traditional3}. However, this approach may miss many potentially effective grasp poses. DexNet2.0 \cite{dexnet2} introduces significant advancements in 3D grasp poses, while GPD \cite{gpd} becomes important for generating 6 degrees of freedom poses in complete space. Subsequently, \cite{graspnet} proposes an end-to-end network that can generate dense 6 degrees of freedom poses for the entire scene, and they also develop a large parallel antipodal grasp dataset~\cite{fang2023robust}. Many studies explore the potential of related parallel antipodal grasping techniques, such as \cite{gou2021rgb, fang2022transcg, wang2021graspness}. As the performance of parallel antipodal grasping approaches that of humans~\cite{fang2022anygrasp}, this work aims to build a multi-finger grasping method based on the parallel antipodal grasping approach of \cite{fang2022anygrasp, wang2021graspness}.

In dynamic scenes, previous works predominantly use a fixed trajectory or select the nearest pose at two adjacent moments as the tracking pose \cite{kim2014catching, menon2014motion, akinola2021dynamic, marturi2019dynamic, morrison2018closing}. In contrast, this paper leverages the circular antipodal representation with the highest similarity between the two frames to generate the tracking pose.

\section{Problem Formulation}
 For a grasp-based multi-finger hand, we can define its pose $\mathcal{G}$ as
\begin{equation}\label{multi-finger definition1}
    \mathcal{G} = [\mathbf{R}\ \mathbf{t}\ \mathbf{B}],
\end{equation}
where $\mathbf{R} \in \mathbb{R}^{3\times3}$ denotes the multi-finger hand rotation, $\mathbf{t} \in \mathbb{R}^{3\times1}$ denotes the translation of multi-finger hand, they are the same as the definition of parallel antipodal gripper (for specific correspondence, please refer to Sec.~\ref{mapping_principles}), and $\mathbf{B}$ denotes the freedom of multi-finger hand. Because multi-finger grasping is a decision problem on a continuous joint space, we simplify the degrees of freedom of the multi-finger hand into $\mathbf{C} \in \mathbb{R}^{c\times1}$ alternative grasp types and $\mathbf{W} \in \mathbb{R}^{w\times1}$ alternative grasp width in this paper. So we can equate Eqn.~\eqref{multi-finger definition1} as
\begin{equation}\label{multi-finger definition2}
    \mathcal{G} = [\mathbf{R}\ \mathbf{t}\ w\ c].
\end{equation}
Let $\mathcal{E}$ denote the environment containing all objects and robots in the scene, $\mathcal{P}$ denote the point cloud taken by the camera from a certain angle and $\mathcal{M}$ denote the representation model, the representation $\mathcal{\hat{R}}$ of scene can be denoted as 
\begin{equation}\label{representation}
    \mathcal{\hat{R}} = \mathcal{M}(\mathcal{P}). 
\end{equation}
Let $q(\mathcal{\hat{R}}, \mathcal{G}, \mathcal{E})$ denotes a binary variable indicating whether an object is successfully grasped or not given grasp pose $\mathcal{G}$ and representation $\mathcal{\hat{R}}$. Our goal is to find a pose $\mathcal{G}_i$ from a set of candidate poses $\mathbf{G}=\left\{\mathcal{G}_1,\mathcal{G}_2,\cdots, \mathcal{G}_n\right\}$ that maximizes the grasp success rate:

\begin{equation}\label{potimizer_target}
    \mathcal{G}_i = \argmax_{\mathcal{G} \in \mathbf{G}} \text{Prob}(q=1| \mathcal{\hat{R}}, \mathcal{G}, \mathcal{E} ).
\end{equation}

\section{Method}

In this section, we will introduce our method in two parts. The first part will cover the circular antipodal representation, and the second part will delve into the detailed design of our algorithm. 

\subsection{Circular Antipodal Representation} \label{antipodal_representaiton}
\begin{figure}[!t]
\centering
\includegraphics[width=0.45\textwidth]{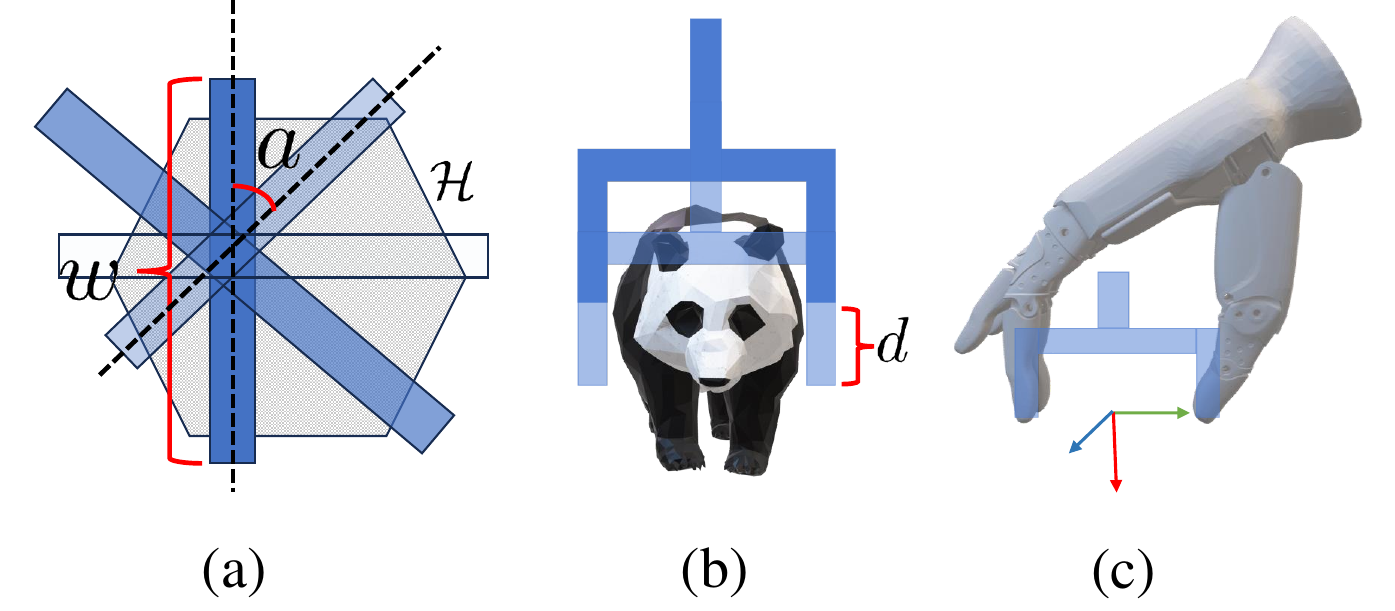}\\
\caption{(a) 2D circular antipodal representations. The depth of color denotes the level of antipodal scores. (b) 3D circular antipodal representations. (c) The coordinate system of parallel antipodal and multi-finger grasping}
\label{fig_representation}
\vspace{-0.5cm}
\end{figure}

We first consider a 2D object. Given a grasp point $\mathcal{X}$, our representation is defined as a set of circular antipodal scores and widths:
\begin{equation}\label{rep}
\mathcal{\hat{R_X}} = \{(s_a,w_a) \mid a=1,2,\cdots,\mathcal{A}\},
\end{equation}
where $s_a$ denotes an antipodal score along the $a$-th in-plane rotation and $w_a$ denotes the mininum antipodal grasp width for the corresponding rotation. The in-plane circular is discretized into $\mathcal{A}$ angles in total. Fig.~\ref{fig_representation}(a) illustrates the antipodal score given a tangent plane $\mathcal{H}$.

Then, for a real-world 3D object, given a grasp point and grasp approach direction, the circular antipodal representation can be defined similarly except that we add an extra depth dimension along the grasp viewpoint:
\begin{equation}\label{rep2}
\mathcal{\hat{R_X}} = \{(s_a^d,w_a^d) \mid a=1,2,\cdots,\mathcal{A}, d=1,2,\cdots,\mathcal{D}\}.
\end{equation}

We discretize the depth dimension along the grasp viewpoint into  $\mathcal{D}$ classes, as illustrated in Fig.~\ref{fig_representation}(b).

\subsection{Algorithm Details}

\subsubsection{Mapping Principles}\label{mapping_principles}

\begin{figure}[!t]
\centering
\includegraphics[width=0.45\textwidth]{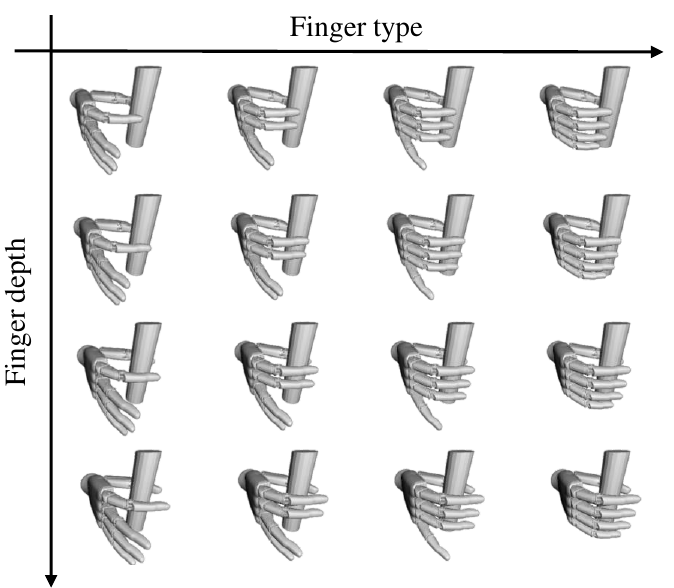}\\
\caption{Multi-finger grasp type, every row denotes that we use the same depth but a different finger type to grasp, and every column denotes the same type but a different grasp depth.}
\label{fig_mapping}
\vspace{-0.5cm}
\end{figure}

To begin with, in order to map the circular antipodal representation to the multi-finger grasp type, we first select the parallel antipodal grasp pose that has the highest antipodal score within the representation. The multi-finger grasp pose is linked to the representation by sharing the same $\mathbf{R}, \mathbf{t}, w$ with the selected parallel antipodal grasp pose. Fig.~\ref{fig_representation}(c) illustrates the coordinate system of both parallel antipodal and multi-finger grasping. For each representation, 16 grasping types are generated, which encompass both precision grasping and power grasping, and cover a wide range of human postures. Fig.~\ref{fig_mapping} illustrates the different grasp types.

\subsubsection{Representation Model} \label{repre_model}
Upon initial examination, the task of obtaining a circular antipodal representation for a partial-view point cloud may seem formidable. However, recent advancements in grasp pose detection algorithms, driven by the application of deep learning techniques, have yielded substantial progress in this regard. These algorithms have demonstrated exceptional proficiency in swiftly and accurately predicting antipodal scores. Noteworthy among these advances are GraspNet-Baseline~\cite{graspnet} and GSNet~\cite{wang2021graspness}, both of which decompose the grasp pose detection challenge into three distinct components: the prediction of grasp point, approach direction, and in-plane antipodal scores and widths along the approach direction. Importantly, the outputs of these algorithms at their final layer inherently provide the circular antipodal representation that is integral to our study. Consequently, for the purposes of this paper, we have opted to employ the readily available GSNet~\cite{wang2021graspness} as our designated representation model.

\subsubsection{Decision Model} \label{model_architecture}

\begin{figure}[!t]
\centering
\includegraphics[width=0.45\textwidth]{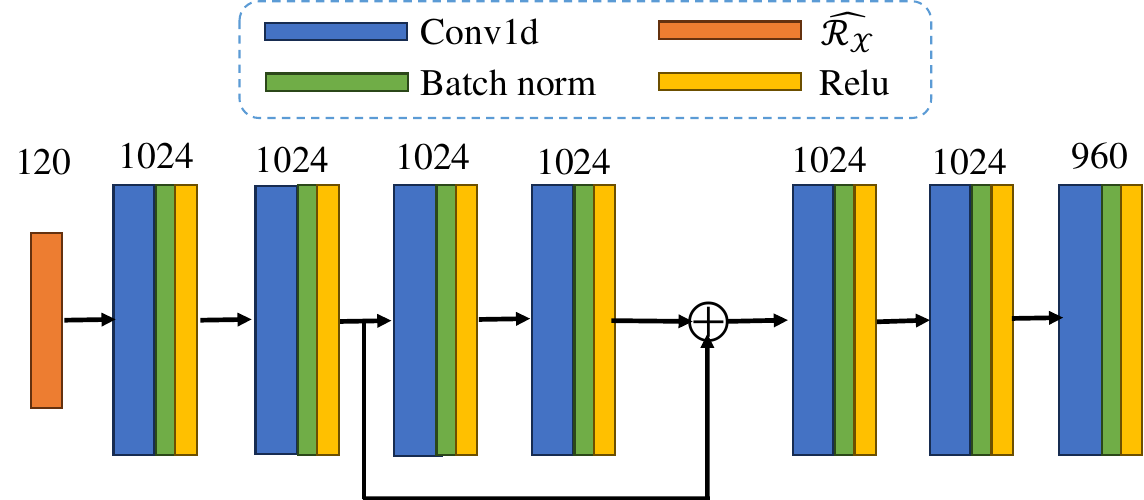}\\
\caption{The architecture of decision model}
\label{fig_model_architecture}
\vspace{-0.5cm}
\end{figure}

The architecture of the decision model is illustrated in Fig.~\ref{fig_model_architecture}. The network comprises seven fully connected layers and a skip connection, which is simple and concise. The input to the model is the circular antipodal representation $\mathcal{\hat{R_X}}$ of each grasp point that is selected. The output of the last layer in the model is a list of scores $O \in \mathbb{R}^{\mathcal{A} \times \mathcal{D} \times \mathbf{C}}$ which denotes the grasp quality score of every multi-finger grasp type in $\mathbf{C}$ and maps to different antipodal scores in the representation $\mathcal{\hat{R_X}}$, where $\mathcal{A}, \mathcal{D}$ and $\mathbf{C}$ are predefined as $12, 5$ and $16$, respectively. The loss function is defined as:
\begin{equation}\label{loss}
L = -\frac{1}{\mathbf{Z}}\sum_{z=1}^\mathbf{Z}y_{c}\log(p_{o,c}).
\end{equation}
Here, $y_{c}$ is the binary indicator (0 or 1) if multi-finger grasp type $c$ successfully grasp an object given $a$ and $d$, $p_{o,c}$ is the predicted probability of $c$ for grasp quality score $o \in O$, and $\mathbf{Z}$ is the batch size. The remaining multi-finger grasp types and different $a, d$ are ignored when calculating the loss because we do not know if they can grasp successfully (Please refer to the data collection and annotation Sec.~\ref{data_collection_and_annotation}). All grasp points share the same decision model parameters. 

\subsubsection{Detection Post-processing}

After obtaining grasp quality scores of each grasp point from the decision model, we select the top 500 multi-finger grasp poses $\mathbf{G}$ according to the grasp quality scores of all grasp points for collision detection and ultimately determine the grasp pose $\mathcal{G}_i$. Among them, collision detection is performed by voxelizing the pre-shaped 3D multi-finger hand model with a voxel size of 0.3 cm, and checking whether there are intersections between the hand voxels and scene point cloud, using Open3D library~\cite{Zhou2018}. We aim to improve the grasp success rate by selecting the grasp quality scores of multi-finger grasp poses with high confidence levels, which are typically above 0.9. We augment the point cloud of a scene by randomly translating or rotating it 10 times to generate additional multi-finger grasp proposals. 

\subsubsection{Dynamic Grasping}
we initialize our approach by selecting a circular antipodal representation $\mathcal{\hat{R_X}}^0$ along with its corresponding multi-finger grasp pose through the utilization of the representation model and decision model for the first frame. Subsequently, we harness the association module within the AnyGrasp framework \cite{fang2022anygrasp} to construct the correspondence matrix between $\mathcal{\hat{R_X}}^0$ and a set of circular antipodal representations $\{\mathcal{\hat{R_X}}_0^1, \mathcal{\hat{R_X}}_1^1, \dots, \mathcal{\hat{R_X}}_n^1\}$ derived from candidate poses $\mathbf{G}$ generated in the subsequent frame. We then select the multi-finger grasp pose which antipodal representation has the highest similarity in the matrix for the next frame. This approach ensures temporal continuity in the context of reactive grasping.

\section{Experiments}

\subsection{Implementation Details}
To evaluate the performance of our multi-finger grasping, we set up several experimental platforms in a real-world environment and conduct our experiments.

For static grasping, we deploy the UR5 robot arm and Intel's RealSense D415 camera to capture the entire scene's point cloud. The camera is positioned overhead in the scene. The gripper is a five-finger dexterous hand manufactured by Inspire\footnote{\url{http://www.inspire-robots.com/product/frwz/}}, which has 6 motors, one for each of the four fingers, and two for the thumb. The hand consists of 12 underactuated joints, with a weight of approximately 0.5 kg. 

For dynamic grasping, we employ the Flexiv Rizon robotic arm, which can update the servo target smoothly. The camera we use is the Intel RealSense L515, which is mounted between the multi-finger hand and the end-link of the robotic arm. We opt for this in-hand setting because images captured from an overhead view would be heavily occluded by the robot when grasping. Additionally, we choose the Intel RealSense L515 camera because it can generate point clouds for objects at close range.

For model training, since we adopt the off-the-shelf GSNet~\cite{wang2021graspness} as the representation model, we only need to train the decision model. The decision model is trained for only 20 epochs, leveraging the Adam optimizer~\cite{kingma2014adam}. The learning rate follows a segmented descent strategy, and the batch size $\mathbf{Z}$ is set to 128 to optimize training efficiency. 

The training and inference of all the models in the following experiments are performed on the Ubuntu 20.04 system, using an Intel i9-10900K CPU and an NVIDIA 2080 Ti GPU. All the code is written in Python language.

\subsection{Data Collection and Annotation}\label{data_collection_and_annotation}

As mentioned earlier, most of the data related to multi-finger grasping in previous works are generated within a simulation environment. Nevertheless, significant deviations may arise due to the inherent differences between the simulation and real environments. Consequently, certain poses may be graspable in simulation, but ungraspable in the real world. For this reason, the multi-finger grasp data obtained from simulation still requires validation within a real environment. So we directly collect data in the real world.

Let us provide an overview of the complete data collection pipeline. Initially, we select around 100 objects ranging in size from $3 \times 3 \times 4$ cm$^3$ to $20 \times 8 \times 15$ cm$^3$ and randomly place some objects within a scene. We then use the representation model to generate dense circular antipodal representations $\mathcal{\hat{R}}$ for the scene, along with randomly selecting a multi-finger grasp type $c$ in Fig.~\ref{fig_mapping} to generate multi-finger grasp pose $\mathcal{G}$ for each grasp point. Following this, we perform collision detection and select the multi-finger grasp pose with the highest antipodal score for actual grasping. During this process, we record whether the grasp is successful, the point cloud of the scene, and relevant information associated with the corresponding grasp point. We collect 5000 grasp attempts in total.

As discussed in Sec.~\ref{mapping_principles}, there is a relationship between the parallel antipodal grasping and multi-finger grasping, such that they share the same $\mathbf{R}, \mathbf{t}, w$. As a result, we can achieve a success rate of over 58\% during data collection. Our primary objective is to identify and address the remaining unsuccessful multi-finger grasps by leveraging the decision model to learn the inconsistencies between the parallel gripper and the five-finger hand.

 We randomly select 500 samples from the entire dataset to create the evaluation dataset, while the remaining samples comprise the training dataset. We construct test scenarios that represent common objects found in daily life, including their shape, material, and texture etc. The data is divided into six categories: hardware, snack, ragdoll, household, toy, and adversarial objects, which contain over 300 objects ranging in size from $2.5 \times 2.5 \times 2.5$ cm$^3$ to $36 \times 4 \times 11.5$ cm$^3$. From the collected data distribution, we observe that deeper and more fingers utilized during grasping increase the likelihood of success but also raise the possibility of collision detection failures.

\subsection{Experimental Results}
\subsubsection{Static Scenes}

\begin{figure}[!t]
\centering
\includegraphics[width=0.45\textwidth]{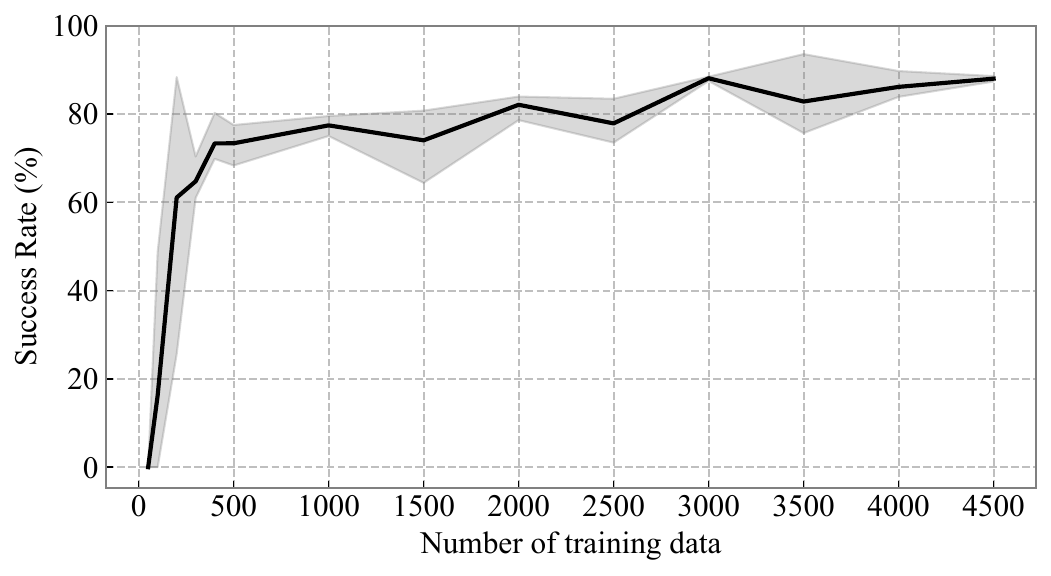}\\
\caption{Performance of different training data on the evaluation dataset}
\vspace{-0.5cm}
\label{fig_test_data_result}
\end{figure}

To evaluate the learning efficiency of our model, we commence with an analysis of its success rate on the evaluation set, utilizing training datasets of varied sizes. The results for each experiment are obtained from three training runs, each consisting of a random subset of the entire training dataset. The results are presented in Fig.~\ref{fig_test_data_result}. We observe that the performance of our model improves as the size of the training dataset increases. When the amount of trial-and-error data exceeds 500, our model achieves a success rate of over 73\%. Furthermore, the average success rate of our model can reach 88\% when using all training samples. 

\begin{figure}[!t]
\centering
\includegraphics[width=0.45\textwidth]{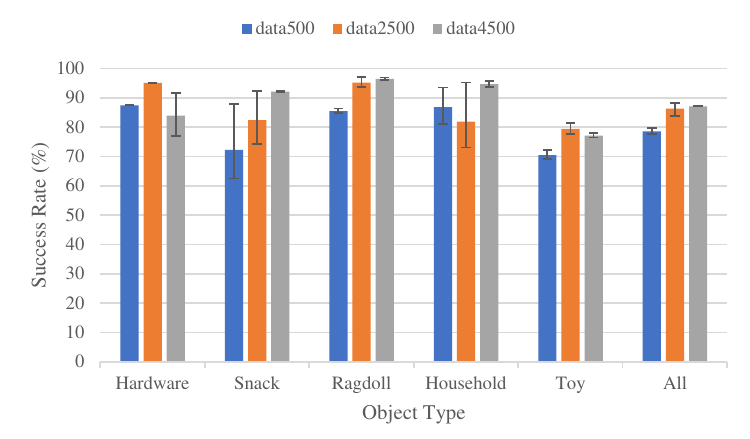}\\
\caption{Performance of different training data on real-world grasping}
\label{fig_evaluation_data_result}

\end{figure}

We then evaluate our method in real-world grasping. Due to the time-consuming nature of real robot grasping, we only test models trained with three different dataset volumes. The overall experimental results are consistent with those obtained from the evaluation dataset and are presented in Fig.~\ref{fig_evaluation_data_result}. The reason for the decrease in the success rate of hardware and toys is attributed to a minor element of randomness in the grasp process. The grasp video can be found in the Appendix (S1-S2, S4-S7).

\begin{figure}[!t]
\centering
\includegraphics[width=0.45\textwidth]{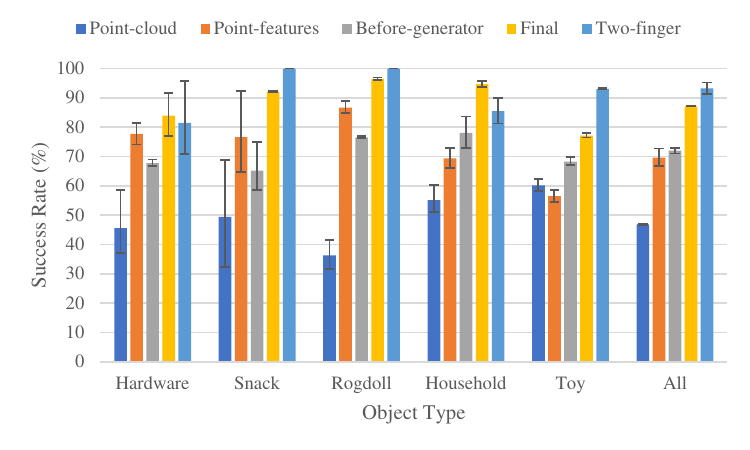}\\
\caption{Performance of different features on real-world grasping}
\label{fig_evaluation_feature_result}

\end{figure}

Next, we evaluate the success rate of using different representations. Models are tested on real-world grasping while trained on the full data set. The results are shown in Fig.~\ref{fig_evaluation_feature_result}. The representation ``Point-cloud'' only uses the Minkowski Engine~\cite{choy20194d} to obtain the point cloud features of the entire scene. The representation ``Point-features'' means the output features of the third last MLP block in the representation model, the ``Before-generator'' is the second last, and the ``Final'' refers to our structured representation. We feed the representation of the grasp point into the decision model in Sec.~\ref{model_architecture} for learning. Our results show that the performance of using the structured circular antipodal representation outperforms other intermediate features by a large margin, with a final success rate of 87\%. The video of the grasping is available in Appendix (S1-S2, S8-S9, S11-S14).

We conduct further evaluation of our model on challenging adversarial objects. The experimental dataset comprises 13 samples from DexNet2.0~\cite{dexnet2} and 49 samples from EGAD~\cite{morrison2020egad}. The results of our model with different trial-and-error numbers and different representation inputs are shown in Fig.~\ref{fig_adversarial}. We can observe from the videos in Appendix (S3, S5, S7, S10, S12, S14) that our method achieves the best grasp performance on adversarial objects when using our structured representation. Notably, we also report the results of grasping with the antipodal grasp pose generated by the representation model, which is denoted as ``Two-finger'' in Fig.~\ref{fig_adversarial}(b). We can see that our multi-finger grasping can outperform its representation model with the aid of an extra decision model.
\begin{figure}[!t]
\centering
\includegraphics[width=0.45\textwidth]{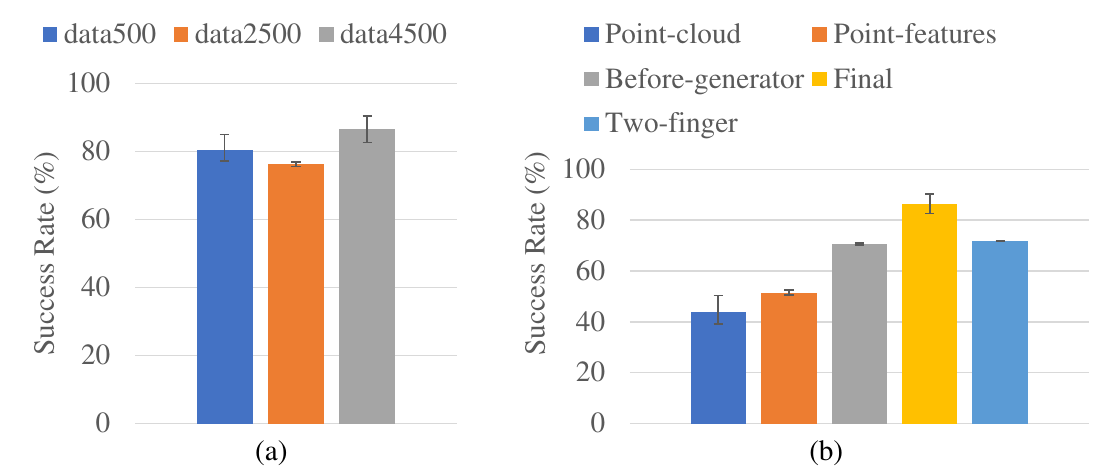}
\caption{(a) Performance of different training data on the adversarial object. (b) Performance of different representations on the adversarial object.}
\label{fig_adversarial}
\vspace{-0.5cm}
\end{figure}

\begin{figure*}[!t]
\centering
\includegraphics[width=0.9\textwidth]{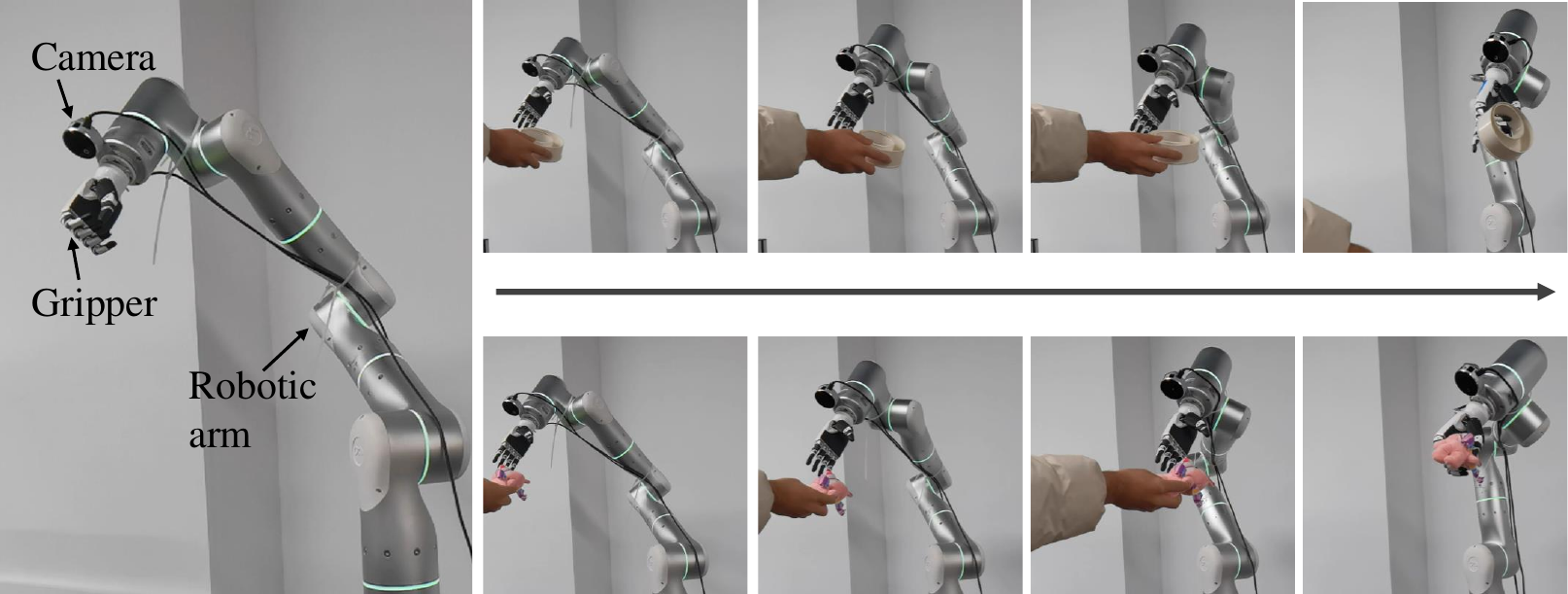}
\caption{Dynamic grasping for hardware and ragdoll, the first column is the initial setting.}
\label{fig_dynamic_grasping}
\vspace{-0.5cm}
\end{figure*}

Finally, we tested the time taken by our system in different phases. The results are shown in Table.~\ref{table_spent_time}. The mean picks per hour (MPPH) of the entire system can reach $167.78$. We can see that both the representation model and decision model run efficiently. The most time-consuming part of our system is collision detection, which costs nearly 20 seconds. The current collision detection runs on the CPU, and we aim to address this problem in the future.

\begin{table}[!t]
\centering
\caption{The time spent in each grasping procedure, RM means the time required by the representation model to generate the representations, DM is the decision model, CD is collision detection, and GAP is the time required by a robot to grasp and place an object.}
\label{table_spent_time}
\begin{tabular}{ccccc}
\hline
\textbf{Grasping Procedure} & RM & DM & CD & GAP\\
\hline
\textbf{Time (s)} & 0.2 & 0.01 & 19 & 2.24\\
\hline
\end{tabular}
\vspace{-0.3cm}
\end{table}

\subsubsection{Dynamic Scenes}

Temporal continuity is crucial in dynamic scenes. An advantage of our framework is that the temporal continuity for multi-finger grasping can be inherited from that of the representation model. Fortunately, the temporal block introduced in~\cite{fang2022anygrasp} is capable of ensuring this. As a result, we use the output of that block as our circular antipodal representations to conduct the handover experiments. We first generate the circular antipodal representations for the first frame and select the best multi-finger grasp pose grasp quality score by the decision model. Subsequently, the temporal block in~\cite{fang2022anygrasp} is used to track the translation and orientation of the representation. The target multi-finger pose is updated according to the translation and orientation of the associated representation.

\begin{table}[!t]
\centering
\caption{The success rate of each category for dynamic grasping.}
\label{table_dynamic_grasping}
\begin{tabular}{cccc}
\hline
\textbf{Category} & Hardware & Snack & Ragdoll\\
\hline
\textbf{Success rate (\%)} & 83.33 & 100.00 & 90.90\\
\hline
\hline
\textbf{Category} & Household & Toy & Adversarial\\
\hline
\textbf{Success rate (\%)} & 71.43 & 90.90 & 83.33\\
\hline
\end{tabular}
\vspace{-0.3cm}
\end{table}

We randomly selected five objects from each category in the static scene and combined them to create the real-world test grasp dataset for the handover experiment. In total, 30 objects were collected. We performed the handover evaluation on the entire real-world grasping twice and achieved a final success rate of 84.51\%. The results for each category are shown in Table.~\ref{table_dynamic_grasping}, and Fig.~\ref{fig_dynamic_grasping} illustrates two samples of dynamic grasping. The video can be found in Appendix~(S16).

\section{Conclusion}

In this paper, we propose an efficient multi-finger grasp method that utilizes supervised learning to grasp objects in complex environments. With a representation model and a decision model, we can generate multi-finger grasp poses. Our training data is collected by trials and errors in the real world that makes it easy to be adapted for different hands. Extensive real robot experiments are conducted to evaluate the effectiveness of our method.

While our proposed multi-finger grasp method has yielded impressive results, we recognize that there is ample room for further improvement. For instance, the multi-finger grasp poses we have defined are not exhaustive, and there are many more poses that human hands can achieve that our method currently cannot replicate. To achieve the same level of dexterity as human hands, we will require more comprehensive datasets, as well as multi-finger hand with greater degrees of freedom. These advancements will enable us to tackle more complex manipulation tasks that build upon the multi-finger hand grasp problem.

{\small
\textbf{APPENDIX}
All experiments related to real-world scenes in this work are recorded in the form of videos. Since the videos are a bit long, we have accelerated a portion of them. All video links can be viewed below.
}
{\small
\begin{itemize}
    \item S1:  \url{https://youtu.be/Rgg-HsM2LNA} 
    \item S2: \url{https://youtu.be/hHFuDTrQh9E} 
    \item S3: \url{https://youtu.be/aWCX-DOfW_o} 
    \item S4: \url{https://youtu.be/fRJQw7jovmc} 
    \item S5: \url{https://youtu.be/eHKvOVoZI6c} 
    \item S6: \url{https://youtu.be/w4zFK75rKXg} 
    \item S7: \url{https://youtu.be/kG0k8HOUFzA} 
    \item S8: \url{https://youtu.be/RZPGgEOBxqM} 
    \item S9: \url{https://youtu.be/QynUVdQdK_0} 
    \item S10: \url{https://youtu.be/cDEcl0D07BM} 
    \item S11: \url{https://youtu.be/4tXBtFnt6Tc} 
    \item S12: \url{https://youtu.be/dfOCi0yN4sY} 
    \item S13: \url{https://youtu.be/WNb5JJykQtQ} 
    \item S14: \url{https://youtu.be/ew6JlynFojc} 
    \item S15: \url{https://youtu.be/8FztVFRcvMY}
    \item S16:\url{https://youtu.be/LdkniuMXYuk}, \url{https://youtu.be/M2VwJjl8FHE}
\end{itemize}
}
{\small
\textbf{ACKNOWLEDGEMENTS}
This work was supported by the National Key Research and Development Project of China (No.2022ZD0160102, No.2021ZD0110704), Shanghai Artificial Intelligence Laboratory, XPLORER PRIZE grants.
}




\printbibliography

\end{document}